\documentclass{article}

\PassOptionsToPackage{numbers, compress}{natbib}

\usepackage[final]{nips}

\usepackage[utf8]{inputenc} 
\usepackage[T1]{fontenc}    
\usepackage{hyperref}       
\usepackage{url}            
\usepackage{booktabs}       
\usepackage{amsfonts}       
\usepackage{nicefrac}       
\usepackage{microtype}      
\usepackage{xcolor}         
\usepackage{times}
\usepackage{latexsym}
\usepackage[T1]{fontenc}
\usepackage[utf8]{inputenc}
\usepackage{microtype}
\usepackage{colortbl}

\usepackage{graphicx}
\usepackage{float}

\newfloat{figtab}{htb}{fgtb}
\makeatletter
  \newcommand\figcaption{\def\@captype{figure}\caption}
  \newcommand\tabcaption{\def\@captype{table}\caption}
\makeatother
\graphicspath{ {./images/} }

\usepackage[table,xcdraw]{}

\usepackage{algorithm}  
\usepackage{algorithmicx}  
\usepackage{algpseudocode}  

\floatname{algorithm}{Algorithm}  

\usepackage{microtype}
\usepackage{booktabs}
\usepackage{wrapfig}
\usepackage{soul}
\usepackage{bbding}
\usepackage{amsmath}
\usepackage{latexsym}
\usepackage{amssymb}
\usepackage{pifont}

\usepackage{multirow} 
\usepackage{makecell}
\usepackage{stfloats}
\usepackage{stackengine}
\usepackage{subfigure}

\usepackage{tabularx}
\usepackage{threeparttable}
\usepackage{hyperref}
\title{MetaXCR: Reinforcement-Based Meta-Transfer Learning for  Cross-Lingual Commonsense Reasoning}

\author{%
  Jie He\\
  School of Informatics\\
  University of Edinburgh\\
  Edinburgh\\
  \texttt{j.he@ed.ac.uk} \\
  \And
  Yu Fu \\
  College of Intelligence and Computing \\
  Tianjin University \\
  Tianjin \\
  \texttt{fuyu\_1998@tju.edu.cn} \\
}

\begin{document}

\maketitle

\begin{abstract}
Commonsense reasoning (CR) has been studied in many pieces of domain and has achieved great progress with the aid of large datasets. Unfortunately, most existing CR datasets are built in English, so most previous work focus on English. Furthermore, as the annotation of commonsense reasoning is costly, it is impossible to build a large dataset for every novel task. Therefore, there are growing appeals for Cross-lingual Low-Resource  Commonsense Reasoning, which aims to leverage diverse existed English datasets to help the model adapt to new cross-lingual target datasets with limited labeled data. In this paper, we propose a multi-source adapter for cross-lingual low-resource Commonsense Reasoning (MetaXCR). In this framework, we first extend meta learning by incorporating multiple training datasets to learn a generalized task adapters across different tasks. Then, we further introduce a reinforcement-based sampling strategy to help the model sample the source task that is the most helpful to the target task. Finally, we introduce two types of cross-lingual meta-adaption methods to enhance the  performance of models on target languages. Extensive experiments demonstrate MetaXCR is superior over state-of-the-arts, while being trained with fewer parameters than other work.
\end{abstract}

\section{Introduction}
Commonsense reasoning (CR) is a basic skill of humans to deal with ordinary situations that involve reasoning about physical and social world \cite{10.1145/2701413}. To endow computers with human-like commonsense reasoning capabilities is hence one of major goals of artificial intelligence. In natural language processing (NLP), different from well-established tasks such as machine translation, natural language generation, question answering, commonsense reasoning interweaves with many other tasks (e.g., conversation generation \cite{ijcai2018-643}, machine reading comprehension \cite{mihaylov-frank-2018-knowledgeable} and exhibits different forms (e.g., question answering \cite{huang-etal-2019-cosmos,talmor-etal-2019-commonsenseqa}, co-reference resolution \cite{Sakaguchi_LeBras_Bhagavatula_Choi_2020,long-webber-2022-facilitating,long-etal-2024-multi}. To facilitate commonsense reasoning research in NLP, a wide variety of CR datasets have been created recently \cite{forbes-etal-2020-social,sap-etal-2019-social}, covering different CR forms and aspects, such as : social interaction \cite{forbes-etal-2020-social,sap-etal-2019-social}, laws of nature \cite{Bisk_Zellers_Lebras_Gao_Choi_2020}, emotional responses \cite{rashkin-etal-2018-event2mind}.

Unfortunately, these datasets are all in English. Leveraging CR resources in the source language of English to enable cross-lingual knowledge transfer to low-resource target languages is therefore desirable for low-resource commonsense reasoning. Among this line of research, training multilingual pretrained language models (e.g., mBERT \cite{devlin-etal-2019-bert}, XLM \cite{conneau-etal-2020-unsupervised}) on the resources of the source language and fine-tuning on the target language data is a promising approach, which has been explored in open-domain QA \cite{xorqa}, semantic role labeling \cite{daza-frank-2020-x}, dialogue \cite{schuster-etal-2019-cross-lingual}. However, using multilingual pretrained language models for cross-lingual CR faces the following two challenges.

\begin{figure}
\setlength{\belowcaptionskip}{-0.4cm} 
\centering
\includegraphics[scale=0.4]{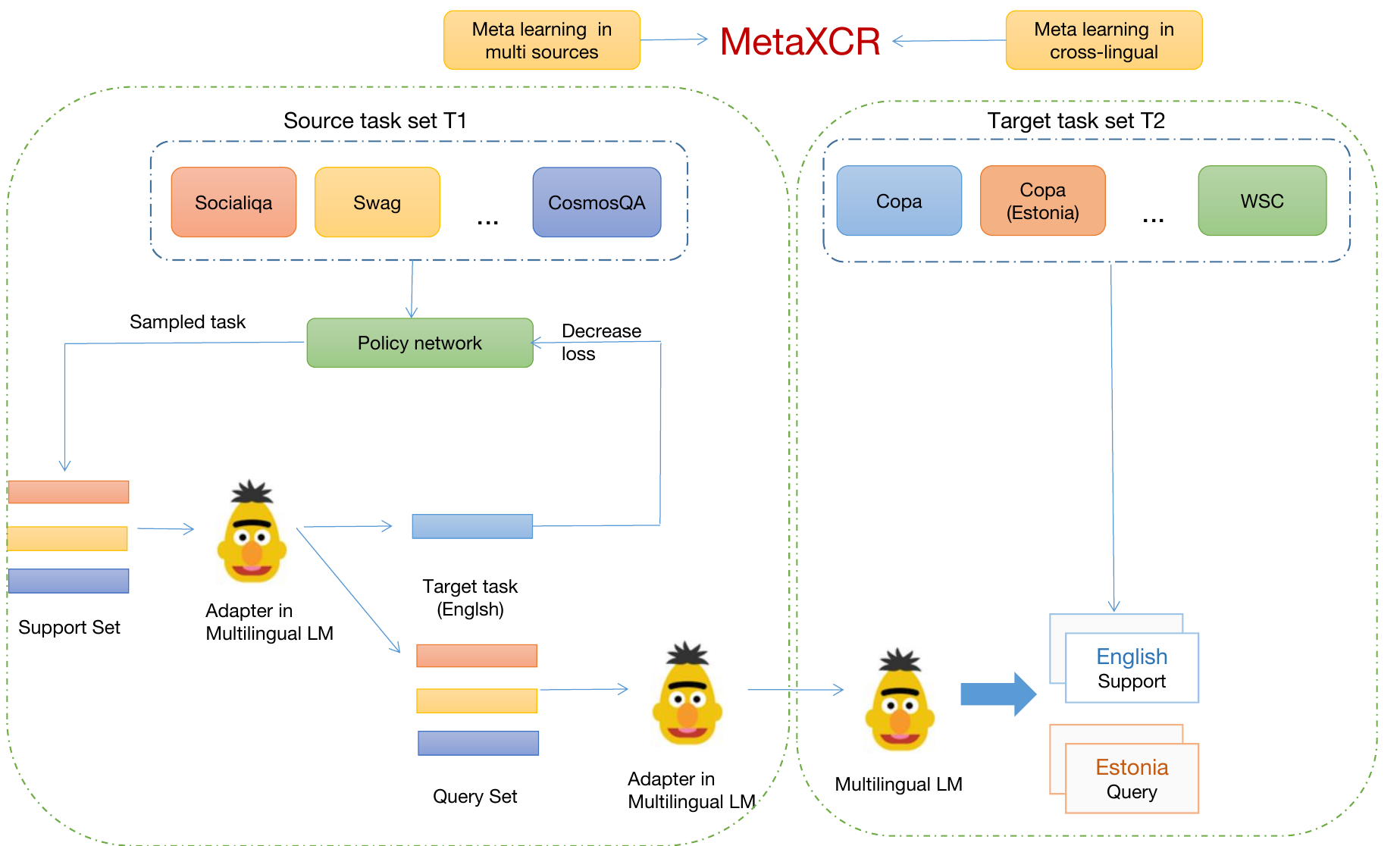}
\caption
 {Our MetaXCR framework.}
\label{method_graph}
\end{figure}

\begin{itemize}
\setlength{\itemsep}{0pt}
\setlength{\parsep}{0pt}
\setlength{\parskip}{0pt}
    \item Language bias in cross-lingual transfer. English-centric resources undoubtedly introduce language bias to commonsense reasoning modeling in other languages \cite{ponti-etal-2020-xcopa}.
    \item Dataset selection in cross-lingual transfer. Previous studies have demonstrated that dataset selection in multi-source transfer learning will have a great impact on the target task, and inappropriate datasets may even hurt performance \cite{bansal-etal-2020-learning,yan-etal-2020-multi-source}.
\end{itemize}

In order to mitigate these issues, we propose a reinforcement-based meta-transfer learning framework for cross-lingual commonsense reasoning (MetaXCR), which performs cross-dataset and cross-lingual transfer learning  to improve the model’s adaptability to the target language dataset. To address the issue of task (i.e., source dataset) sampling, we propose a reinforcement-based learning algorithm for adaptively sampling CR tasks of the source language, which are conducive to the target language CR task. Specifically, as shown in Figure \ref{method_graph}, we first randomly sample a batch of source tasks to form a support set for pseudo-update of the adapter, and then sample a batch of instances from the target CR task to form a query set to meta-update the adapter. In the sampling process, we train an LSTM network to predict the sampling probability of the chosen source tasks via the proposed reinforcement learning algorithm. Using the loss decrease of the pseudo model on the target task as reward to the policy network, we enable the model to automatically learn a strategy that is beneficial to its sampling. For the source to target language adaptation, we sample from the source dataset corresponding to the target task to create a support set, and from the dataset of the target task to build a query set, which are used to update parameters of modules of MetaXCR other than the adapter.

To summarize, our contributions are three-fold:
\begin{itemize}
\setlength{\itemsep}{0pt}
\setlength{\parsep}{0pt}
\setlength{\parskip}{0pt}
\item We propose MetaXCR, to the best of our knowledge, the first work that uses meta-learning and reinforcement learning to for low-resource commonsense reasoning.

\item We propose adapters to solve the cross-lingual and cross-dataset challenges, which enables the model to explore the interaction and combination of different CR tasks and languages.

We propose a reinforcement learning algorithm for sampling source tasks in meta-transfer learning of commonsense reasoning and compare with other sampling strategies. 

\item Experiments show that our MetaXCR has achieved state-of-the-art on cross-lingual low-resource commonsense reasoning datasets, XCOPA \cite{ponti-etal-2020-xcopa}, only requiring 1.6\% of the training parameters.
\end{itemize}
\section{Related work}
\textbf{Meta learning}

Meta-learning, which is ``learn to learn'', trying to enhance the generalization ability of models and dealing with the problem of rapid adaptation to low-resource tasks, is widely used in computer vision \cite{Guo_2020_CVPR,Jamal_2019_CVPR,Liu_2020_CVPR,rajasegaran2020itaml,Wang_2020_CVPR,he-etal-2022-evaluating,long2024leveraginghierarchicalprototypesverbalizer}. Recent years have gained an increasing attention of meta-learning in NLP. 
\citet{yan-etal-2020-multi-source} propose a meta-learning method on the monolingual MCQA task to integrate multiple question answering tasks for low-resource QA. Unfortunately, they ignore correlations among different source tasks,and fine-tuning the entire model is not suitable for low-resource tasks as it is prone to overfitting. \citet{DBLP:journals/corr/abs-2012-11896} propose an approach to improve sampling in the meta-learning process. Different from their work where the goal is to train a generalization model, we are  interested in exploring  related tasks to train a better model for the target task. \citet{DBLP:journals/corr/abs-2102-09397} also  uses adapters to perform meta-training on summarization data from different corpora. However, they use rule-based sampling method to pre-select tasks for meta-training, and rules are task-specific. By contrast, our reinforcement-based sampling method can be applied to any tasks. To sum up, previous works in this aspect either use meta-learning to cross-lingual issues for the same task or to generalize over different tasks for the same language. While we propose a meta-learning framework for solving both cross-lingual and cross-task challenges in commonsense reasoning  simultaneously. 

\textbf{Commonsense reasoning}

Recent years have witnessed that many commonsense reasoning data sets have been proposed. \citet{gordon-etal-2012-semeval} create COPA for causal inference while  \citet{rahman-ng-2012-resolving} present Winogrand Scheme Challenge (WSC), a dataset testing commonsense reasoning in the form of anaphora resolution. Since the size of these datasets is too small, effective training can not be obtained until the recent emergence of pre-training methods has brought new improvements to these datasets \cite{he-etal-2019-hybrid}. Meanwhile, large-scale commonsense reasoning datasets have been curated  \cite{Bisk_Zellers_Lebras_Gao_Choi_2020,huang-etal-2019-cosmos,lin-etal-2020-commongen,Sakaguchi_LeBras_Bhagavatula_Choi_2020,sap-etal-2019-social,talmor-etal-2019-commonsenseqa,long-etal-2020-ted,long-etal-2020-shallow}  to train neural CR models. 
More challenging CR datasets have also been constructed, e.g., HellaSwag \cite{zellers-etal-2019-hellaswag} from Swag \cite{zellers-etal-2018-swag} via adversarial filtering, as powerful language models have either established new state-of-the-art performance \cite{NEURIPS2020_1457c0d6}.
RAINBOW \cite{DBLP:journals/corr/abs-2103-13009}, which uses multitask learning to provide a pre-trained commonsense reasoning model on top of various large-scale commonsense reasoning datasets, is similar to our work. However, the RAINBOW  benchmark is evaluated only on  English commonsense reasoning datasets while we focus on cross-lingual low-resource commonsense reasoning scenarios.

\section{MetaXCR}
	\begin{algorithm} 
	\small
		\caption{MetaXCR}  
		\begin{algorithmic}[1] 
			\Require Task distribution over source $\rm p(T_s)$; 
			         data distribution over target $\rm p(T_t)$;
			         adapter $\rm A$ with parameters $\rm \theta_{A}$;
			         pretrained learner $\rm B$ with parameters $\rm \theta_B$;
			         policy network with parameters $\rm \phi$
			         learning rates in XMSMS
			         $\rm \alpha_{ctml}$,$\rm \alpha_{clml}$,$\rm \beta_{ctml}$,$\rm \beta_{clml}$,$\rm \gamma$
			 \State freeze the pretrained learner parameters $\rm \theta_{B}$
			\While{not done}  
			\State  Generate k-dim vertor of sampling probabilities $\rm P_{T}=(P_{T_{s_1}},P_{T_{s_2}},...,P_{T_{s_k}})$ using $\rm f(\phi)$ 
			\State  choose M source task set $\rm \{T_{s_j}\}_{j=1}^M$ according to Eq.  (\ref{rl_estimate}) 
			\State  Sample task $\rm T_{s_j}^i$ from  $\rm p(T_{s_j})$, get $\rm \{T_{s_j}^i\}_{j=1}^M$, $\rm s_j\in\{1,...,k\}$ for meta-learning
		    \For {all $\rm \{T_{s_j}^i\}_{j=1}^{M}$}
		    \State using support set $\rm D_{support}^{s_j}$ in $\rm T_{s_j}^i$ and compute adapted parameters:
		    \State $\rm \theta_{A}^{s_j}=\theta_{A}-\alpha_{ctml}\nabla_{\theta_{A}} L_{D_{support}^{s_j}}(f(\theta_{A}))$
		    \State using query set $\rm D_{query}^{s_j}$  in $\rm T_{s_j}^i$ and compute the update loss: 
		    \State  $\rm L_{s_{j}} = L_{D_{query}^{s_j}}(f(\theta_{A}^{s_j}))$
		    \EndFor
		    \State Update $\rm \theta_{A} = \theta_{A}-\beta_{ctml}\nabla_{\theta_{A}}\sum_{j=1}^M L_{s_j}$
		    \EndWhile
		    \State
		    
		    \State unfreeze the pretrained learner parameters $\rm \theta_{B}$ and freeze the trained parameters $\rm \theta_{A}$
		    \For{all language $\rm \ell$ }
		    
		    \While{not done}  
		    \State   Sample target english task set $\rm T_{t_e}^i \sim p(T_{t_e})$ 
			\State   Sample target language task set $\rm T_{t_{\ell}}^i \sim p(T_{t_{\ell}})$  
		   
		    \State using support set $\rm D_{support}^{t_e}$  in $\rm T_{t_{e}}^{i}$ and compute adapted parameters:
		    \State $\rm \theta_{B}^{t_{\ell}} = \theta_{B}-\alpha_{clml}\nabla_{\theta_{B}} L_{D_{support}^{t_e}}(f(\theta_{B}))$ 
		    \State using query set $\rm D_{query}^{t_{\ell}}$ and compute the update loss:
		    \State $\rm L_{t_{\ell}} =  L_{D^{t_{\ell}}_{query}}(f(\theta_{B}^{t_{\ell}}))$
		    \State Update $\rm \theta_{B} =  \theta_{B}-\beta_{clml}\nabla_{\theta_B} L_{t_{\ell}}$
		    \EndWhile
		    \EndFor
		\end{algorithmic} 
	\label{algorithm1}
	\end{algorithm}  
	
\textbf{Problem Formulation:} \textbf{\textit{Cross-lingual Cross-task Low-Resource commonsense reasoning}} is a task that requires a model to solve CR problems where the training data is not available or sufficient and the language of the target task is different from the language of source tasks. Let`s denote the labelled training data of source  tasks as  $\rm T_s=\{\{T_1^i\}_{i=1}^{n_1},...,\{T_k^i\}_{i=1}^{n_k}\}$, where $\rm T_{k}=\{T_{k}^i\}_{i=1}^{n_k}$ represents all tasks in the $\rm k$-th source dataset. The training data of source datasets in this paper are all in English. The target evaluation tasks are $\rm T_t=\{\{T_1^i\}_{i=1}^{n_1},...,\{T_{\ell}^i\}_{i=1}^{n_{\ell}}\}$, where $\rm T_{\ell}=\{T_{\ell}^i\}_{i=1}^{n_{\ell}}$ denotes the target task in the $\rm \ell$-th target language. For a single tasks in source dataset $\rm s_j \in \{1, \dots, k\}$ denote as $\rm T_{s_j}^{i}$, it is composed by support batch denote as $\rm D_{support}^{s_j}$ and query batch denote as $\rm D_{query}^{s_j}$. The target evaluation tasks are in CR forms different from those of the source tasks (e.g.,  Swag vs. WSC). As our goal is to deal with low-resource CR,  the  target evaluation tasks can be divided into two scenarios:  zero-shot and few-shot. Few-shot means  $\rm T_{\ell}$ has labelled instances that can be used for further fine-tuning. \textbf{Zero-shot} indicates that there are no labeled instances for $\rm T_{\ell}$ at all and that the model trained on the source tasks will be directly evaluated on the corresponding target task. Specially, We use $\rm f(\theta)$ to denote the trained multilingual model. 

\textbf{General introduction to our method:} As shown in Figure \ref{method_graph}, we first sample numbers of tasks $\rm \{ T_{s_{j}}^{i} \}_{j=1}^{M}$ from the source task set $\rm T_{s}$. The data of the sampled tasks are then used to update the adapter parameters via the proposed  meta-learning described in section \ref{sctml}. As different source tasks have different contributions to the target task, we propose different  sampling strategies choose desirable source tasks for the target task in the cross-task meta-learning process  of our MetaXCR in section 3.2.3.
As source tasks are all in English while our target tasks are in a range of different languages, we introduce cross-lingual meta-learning in section \ref{sclml}. 
In the following, we first introduce the proposed commonsense reasoning  framework, and then elaborate on the cross-lingual and cross-task meta-transfer learning process and the reinforcement-based sampling strategy. 
\subsection{Commonsense Reasoning Framework}
\textbf{Base Model} We choose the state-of-the-art pretrained multilingual  BERT (mBERT) \cite{devlin-etal-2019-bert} as the base commonsense reasoning model.  Special token  [CLS]  is  added at the beginning of each sentence to aggregate information. As multiple–choice selection  tasks  differ  in  the  numbers  of  candidate answers  (e.g.,2 candidate answers per question  in COPA vs. 5  in CommonsenseQA), a classifier with a fixed number of classes is not a good fit for this scenario. We thus follow \citet{sap-etal-2019-social}'s method . First, we concatenate  the  context,  question,  and answer, using  specific separator tokens. For BERT, the concatenation is formatted as $\rm [CLS] <context> [UNUSED] <question> [SEP] <answer_i> [SEP]$. Then, we use a multilayer perceptron to compute a score $\rm \hat y_i$ with  the classifier token $\rm h_{CLS}\epsilon R^H$: 
$$\rm \hat y_i=W_2tanh(W_1h_{CLS}+b_1) \qquad (W_1\epsilon R^{H \times H},b_1\epsilon R^H, W_2\epsilon R^{1\times H})$$  Finally, we estimate the probability distribution over candidate answers using a softmax layer: $\rm \hat y =softmax([\hat y_1,...,\hat y_N])$.  The final  answer predicted by the base model corresponds to the triple with the highest probability.

\textbf{Adapters} Inspired by \citet{pfeiffer-etal-2020-mad}, we use task-related adapters in mBERT  to separate language and tasks to achieve better performance on cross-lingual  CR.  For each layer $\rm l$ in mBERT, we insert an adapter module which is a bottlenecked feed-forward network consisting of a down-layer $\rm f_{\theta_{d}}$ and an upproject layer $\rm f_{\theta_{u}}$. A skip-connection from input to output is also established. The adapter (AP) can be formulated as:
\begin{equation}
\rm 
    AP(h)=f_{\theta_{u}}(ReLU(f_{\theta_{d}}(h))+h)
\end{equation}
Only parameters in adapters  are updated during fine-tuning on a downstream task, which  aim to capture knowledge that is task-specific but generalises across languages.
\subsection{Cross-Lingual and Cross-Task Meta-Transfer Learning}
\subsubsection{Cross-Task Meta-Learning}
\label{sctml}
CTML is presented  in lines 1-13 in Algorithm \ref{algorithm1}. In CTML, for each meta-training iteration, we sample $\rm L$ source dataset. For each sampled source dataset, we need to randomly sample instances from it to obatin support and query batch to form a task denoted as {$\rm T_{s_1},T_{s_2},...,T_{s_L}$} to construct  meta-train tasks in meta learning. The learning rate for the inner and outer loop are different: $\rm \alpha$ denotes the learning rate for the inner loop, while $\rm \beta$ for the outer loop. The inner loop aims to learn the task information from support batch. For a task $s_{i} \in \{ s_1, \dots, s_L\}$, The task specific  parameters $\rm \theta^{s_{i}}$ are updated by 
\begin{equation}
    \theta^{s_{i}}=\theta-\alpha\nabla_\theta L_{D_{support}^{s_i}}(f(\theta))
    \label{meta_support}
\end{equation}
where $\rm \nabla_\theta L_{D_{support}^{s_i}}(f(\theta))$ is computed as the gradient of  the cost function $\rm L_{D_{support}^{s_i}}(f(\theta))$ with respect to model parameter $\rm \theta$. 

In the outer loop, $\rm L_{D_{query}^{s_i}}(f(\theta^{s_i}))$ is calculated with respect to $\rm \theta^{s_i}$, to update the meta model on the corresponding query batch $\rm D_{query}^{s_i}$.

It is worth noting that $\rm f(\theta^{s_i})$ is an implicit function of $\rm \theta$. As the second-order Hessian gradient matrix requires  expensive computation, we employ a first-order approximation (FOMAML), which ignores second-order derivatives:
\begin{equation}
   \theta=\theta-\beta\frac{1}{L}\sum_{i=1}^{L}\nabla_\theta L_{D_{query}^{s_i}}(f(\theta^{s_{i}}))
   \label{meta_query}
\end{equation}
We keep running the CTML module until the  meta-model  converges.   After performing CTML, the CLML module will be applied upon the learnt meta representations for the final transfer learning on the target language.
\subsubsection{Cross-Lingual Meta-Learning}
\label{sclml}
CLML is formulated in  lines  15-26 in Algorithm \ref{algorithm1}.  Following MAML \cite{pmlr-v70-finn17a}, in each  iteration, we sample  a support and query batch from a same task, but with the source and target language respectively. In order to make representations learned before well adapted to different target languages, we sample labelled instances  for the support set of tasks in CLML from the training data of source tasks in the source language (i.e., English in our experiments). For the query set in CLML, we sample from the target task data $\rm T_{t_{\ell}}$ in the target language. The entire procedure  of CLML is similar to CTML. For each target language, we use  support batches from the source language  for  the  inner-loop update and  query batches from the target language for  the outer-loop update. In addition, since only the task adapter is trained  in CTML, task representations can be learned. Therefore, in CLML, we fix learned  task representations and only train mBERT itself. In order to prevent overfitting, we only train the last layer of mBERT. 
\subsection{Target-adapted meta sampling}
For MAML, it is crucial to properly define meta-train tasks. An intuitive strategy is to use tasks from all datasets during every meta-training step. However, this is not a good choice since different source dataset might have different data distributions. A single undesirable task in a dataset might optimize the model towards a wrong direction. We therefore propose a reinforced strategy to choose source tasks for meta-training. In our experiments, we compared the proposed sampling strategy to other heuristic sampling strategies.

\subsubsection{Heuristic Sampling and Motivation for Reinforced Sampling}
For each MAML step, we sample data to form a task $\rm  T_{s_{j}}^{i}$ for every source dataset $\rm s_{j}$. Our goal is to train good initialization parameters for the multilingual model on the given target dataset.We can calculate a general loss $\rm L_{original}$ on the original model $\rm f(\theta)$ using sampled data from the target dataset. After optimizing the $\rm f(\theta)$ with $\rm D_{support}^{s_{j}}$ of $\rm s_{j}$ according to Eq. (\ref{meta_support}), we can get task-specific model $f(\theta^{s_{j}})$ and get task-specific loss $\rm L_{s_{j}}$ using the same sampled data with general loss.

To achieve the goal, we can use the countdown of the $\rm L_{s_{j}}$ as a guiding signal to sample a corresponding dataset and can be formulated as:
\begin{equation}
    \rm P_{i}^{s_{j}} = \frac{D_{s_{j}}}{ \sum_{j=1}^{k}{D_{s_{j}}}}
\end{equation}
where $\rm s_{j}$ denote the task from the $\rm j$-th source dataset, $\rm i$ denotes the MAML step. $\rm D_{s_{j}}$ is the countdown of $\rm L_{s_{j}}$ which can be seen as how good the model might be after being trained on the chosen task. 

A simple way is to just use the $\rm P_{i}^{s_{j}}$ as the probability to choose the corresponding task to update the model through Eq. (\ref{meta_query}). However, this heuristic sampling strategy might not consistent with the distribution of the entire source dataset and ignore the long-range dependencies among a sequence of source dataset sampling decisions. Inspired by  \cite{xiao2020adversarial}, we use a LSTM-based network to catch the long-term sampling dependencies to form the meta-training tasks. Since we do not have any annotated labels to train the LSTM-based network, we use REINFORCE, a policy gradient algorithm, for our proposed reinforcement-based meta sampling. 
\subsubsection{RL algorithm}
Let $\rm f_\phi(\cdot)$ denote the network trained by reinforcement learning, and $\rm \phi$ be parameters to be tuned. For the REINFORCE algorithm, the gradient can be calculated as:
\begin{equation}
\rm
    \nabla_{\phi}J(\phi)  = E_{\tau \sim f_{\phi}(\cdot)}[R(\tau)  \nabla_{\phi} log f_{\phi}(i)]
\end{equation}
where $\rm \tau$ is the trajectory and $\rm R(\tau)$ is the total reward of the trajectory. $\rm i$ denotes action in the trajectory $\rm \tau$. During REINFORCE training, we use the reward of the total trajectory to guide how to update the network.

 we use a baseline to reduce the high variance of the traditional REINFORCE algorithm and let the reward not always be non-negative. To get the expectation, we sample limited numbers of trajectories and use the sampled trajectories to estimate the gradient:
 \begin{equation}
     \rm \nabla_{\phi}J(\phi) \approx  \frac{1}{N} \sum_{n=1}^{N} \sum_{g=1}^{G} \nabla_{\phi}(R(\tau^{n}) - b))log f_{\phi}(i_{g}^{n})
\label{rl_estimate}
 \end{equation}
 where $\rm b$ is the baseline value, $\rm \tau^n$ is the  $\rm n$-th sampled trajectory, $\rm i_{g}^{n}$  denote the $\rm g$-th action in the $\rm n$-th trajectory.

\subsubsection{Our Sampling Approach}
We use the same network architecture \cite{xiao2020adversarial} together with a feedforward and attention layer. For REINFORCE training, We use the $\rm L_{original}$ as the baseline, and $\rm (L_{orignal} - L_{s_j})$ as the reward of choose the $\rm j$-th source dataset in the MAML steps denote as $\rm D_{j}$. This reward not only fit our motivation to guide the MAML toward a good initialization point for target dataset but also incorporate the baseline in the final reward. At every MAML step $\rm s$, we feed the $\rm (D^{s-1}_{1}, D_{2}^{s-1}, \dots, D_{k}^{s-1})$ into the policy network together with the probabilities $\rm (P_{1}^{s-1}, P_{2}^{s-1}, \dots, P_{k}^{s-1})$, which is estimated in the previous step of the policy network. Then we can obtain a probability distribution over source datasets denoted as $\rm (P_{1}^{s}, P_{2}^{s}, \dots, P_{k}^{s})$. For training the feed-forward and attention layer, we calculate every source datasets reward $\rm \{D_{j} \}_{j=1}^{k}$at every step.

As showed in Eq. (\ref{rl_estimate}), we use the sampled trajectories to get the gradient to update the policy network. It might fast converage to a local minmum and the policy will become a deterministic policy. To alleviate this issue, we incorporate $\rm \epsilon-greedy$ technique into the sampling process as \cite{dong2018banditsum}.

It regards sampling process as a progressive process and sampling in the front affects the probability of sampling in the back. The most important part is to incorporate the $\rm \epsilon-greedy$ technique into the sampling process, to calculate the log of trajectory  which is required in Eq. (\ref{rl_estimate}):
\begin{equation}
\rm 
    \log f_{\phi}(i^{n}) = \log \big(\prod_{g=1}^{G}(\frac{\epsilon}{k - g + 1} +\frac{(1 - \epsilon) * P_{i_{g}}^{s}}{\sum_{g=1}^{G} P_{i_{g}}^{s}  - \sum_{z=1}^{g-1} P_{i_{z}}^{s}} )\big)
\end{equation}
By setting  $\rm \epsilon$, we can control dataset sampling probability estimation. 

The $\rm \epsilon-greedy$ techniques can be seen as the trade-off between exploration and exploitation. Large $\rm \epsilon$ indicate a high probability towards randomly sampling, which results in a high exploration rate. 

\section{Experiments}
\subsection{Datasets}
\subsubsection{Source Datasets}
We used six commonsense reasoning datasets as our source datasets. The detail description can be found in Appendix A.
\subsubsection{Target Dataset}

    We used XCOPA as our target dataset.  XCOPA \cite{ponti-etal-2020-xcopa} comprises 100 annotated instances in the validation set and 500 instances in the test set, which are translations in multiple target languages  from the  English COPA validation and test set.  COPA \cite{gordon-etal-2012-semeval} is a dataset that aims to benchmark causal reasoning in a  binary classification setting. Since  Haitian Creole (HT) and   Quechua (QU) are not covered by mBERT, we didn't report results on these two languages.
    
\subsection{Baselines}
We compared our method against the following baselines.
\begin{itemize}
\setlength{\itemsep}{0pt}
\setlength{\parsep}{0pt}
\setlength{\parskip}{0pt}
    \item \textbf{Sequential fine-tuning}: This baseline  sequentially updates model parameters by fine-tuning models  on each task. The  order of tasks for  fine-tuning is predefined  and at each step the model is initialized with  parameters learned from the previous fine-tuning.
    \item \textbf{Multi-task learning}: All tasks are trained simultaneously in hoping that  shared  representations across tasks are learned to enable  the model to generalize to the target task.
    \item \textbf{M-BERT}: This comes from \citet{ponti-etal-2020-xcopa}, which use soicliaqa to fine-tune mBERT, and then continue to  fine-tune mBERT on the XCOPA data of the target language.
    \item \textbf{Mad-X$^{mbert}$}: This comes from \citet{pfeiffer-etal-2020-mad}, a state-of-the-art model on XCOPA. It uses  a task adapter and a language adapter in each layer of BERT, and uses socialiqa to train the task adapter, and  texts in different languages to train the language adapter through the masking method. Here we reproduce their model with the original mBERT as the backbone. 
\end{itemize}
\begin{table}[]
\setlength{\belowcaptionskip}{-0.4cm}
\label{label1}
\small
    \centering
    \begin{tabular}{lc|ccccccccc|c}
    \bottomrule
    
         Methods& en&et&id&it&sw&ta&th&tr&vi&zh&avg \\ 
        \hline \rowcolor{gray!40}
         \multicolumn{12}{c}{Zero-Shot Learning}\\
         \hline
         M-BERT& 62.2&55.2&57&57&51&52.2&51&53.2&59.2&64.4&56.24\\
         MAD-X$^{mbert}$ &60.6&50.6&-&-&-&-&-&\textbf{58.2}&58.2&\textbf{64.8}&-\\
         Seq&61.6&52.2&57.6&54.4&\textbf{53.4}&52.6&52.2&51.8&58&63.8&55.76\\
         Multi&65.2&53&57&\textbf{58}&\textbf{53.4}&54&\textbf{53.8}&53&56.2&61&56.46\\
         Ours&\textbf{66.6}&\textbf{54.6}&\textbf{59}&54.2&49.8&\textbf{54.6}&\textbf{53.8}&54&\textbf{60.4}&60.8&\textbf{56.78}\\
        \hline \rowcolor{gray!40}
         \multicolumn{12}{c}{Few-Shot Learning}\\
        M-BERT &63.2&52.2&58.2&57.2&51&\textbf{57.2}&52.6&54.6&57.8&52.4&55.64\\
        MAD-X$^{mbert}$ &65.4&51.2&-&-&-&-&-&57&\textbf{60.2}&63.4&-\\
        Seq&64&52&55.4&56.8&\textbf{52}&51.2&51.2&\textbf{58.6}&58.4&62&56.16\\
        Multi&62.08&49.6&56.8&\textbf{60.8}&50.6&54.2&53&51.4&55&61.4&55.56\\
        Ours&\textbf{68}&\textbf{54.4}&\textbf{59.8}&56.6&50.6&55.2&\textbf{54.6}&53.6&58.8&\textbf{64.2}&\textbf{57.58}\\
         \toprule
    \end{tabular}
    \caption{Accuray results on the XCOPA test set for zero- and few-shot MetaXCR for different target languages. }

    \label{tab:my_label}
\end{table}
\section{Results}
\textbf{Zero-Shot Learning:} We carried out experiments with  MetaXCR in  a zero-shot setting, where  only CTML was performed as introduced in section 3.2.1. We report the impact of CTML for each target language as a difference in accuracy with and without meta-learning on top of the baseline method on the test set (Table \ref{label1}). Overall, we observe that our zero-shot  MetaXCR outperforms the baselines and results reported by \citet{ponti-etal-2020-xcopa}. It establishes new  state-of-the-art performance on several different target languages for zero-shot cross-lingual CR. 

\textbf{Few-Shot Learning}: For few-shot   learning, we used the checkpoint model of MetaXCR after CTML was trained and then fine-tuned the checkpoint model  on  the  development  set  (400 instances) of corresponding target languages. The fine-tuned model was finally evaluated  on the test set.  Few-shot learning results in Table \ref{tab:my_label} demonstrate that  our method also achieves  state-of-the-art results. Compared with the zero-shot setting, MetaXCR obtains  significant  improvements on 7 languages (except et, sw, ta) ( from +0.8\% to +3.4\%)  over M-BERT.  Additionally, the performance of M-BERT, sequential fine-tuning and multi-task learning has dropped on most languages after few-shot learning. In contrast, our few-shot MetaXCR is better than zero-shot MetaXCR, suggesting that it benefits from the proposed cross-lingual meta-learning method.

\section{Analyses }

\subsection{Single-task vs Multi-task Adapter}

We use multiple source tasks for cross-task meta-transfer learning. We therefore want to know the effects of each task (e.g., Swag, CosmosQA etc.)  and their different combinations on the target task (i.e., COPA). We conducted experiments to evaluate these effects, which are shown in Table \ref{compare_adapter}. Clearly, using multiple source tasks is much better than using single source task. 
\begin{wraptable}{r}{4cm}

    \centering
    \scalebox{0.7}{
    \begin{tabular}{|l|cc|}
    \bottomrule
        \multirow{2}{*}{Tasks}&\multicolumn{2}{c|}{COPA}\\
        &Dev&Test\\
        \hline
         $\alpha$NLI&58.2&56.4  \\
         CommonsesenQA&63&61.6\\
         CosmosQA&59.8&62.4\\
         Hellaswag&55.6&53.8\\
         SocialIQA&62.2&63.8\\
         Swag&61.4&55.6\\
         \hline
         Three&64.4&65.2\\
         Four&68&65.8\\
         Five&66.6&64.2\\
         Six&\textbf{68.8}&\textbf{66.6}\\
         \toprule
    \end{tabular}
    }
    \caption{Single-task adapter vs.  multi-task adapter on COPA. For experiments with multiple source tasks, we randomly select $\rm n$ ($\rm  n$ = 3-6) tasks from all source datasets. }
    \label{compare_adapter}
\end{wraptable}

\subsection{Sampling Strategy}

            


\begin{figure*}[tb] 
\setlength{\belowcaptionskip}{-0.2cm} 
  \begin{minipage}{0.55\textwidth} 
    \centering 
\includegraphics[scale=0.5]{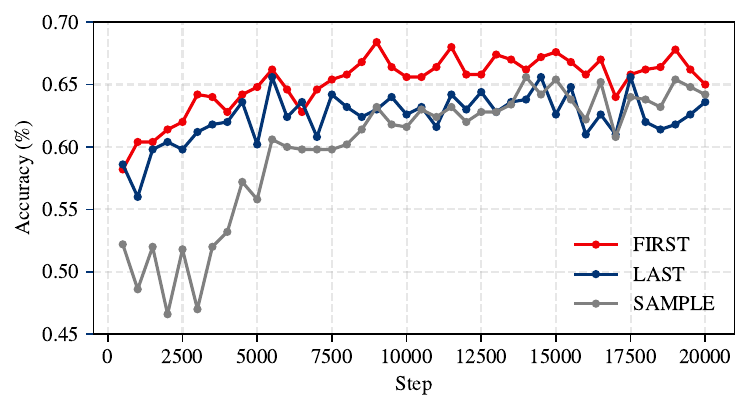}
    \caption
{The test accuracy curve of the COPA for \\ \hspace{2em}three sampling strategries
}
    \label{test_acc} 
  \end{minipage}%
\makeatletter\def\@captype{table}\makeatother
  \begin{minipage}{0.55\textwidth}
    \centering
            \begin{tabular}[width=5\linewidth]{|l|cc|}
            
    \bottomrule
        \multirow{2}{*}{Tasks}&\multicolumn{2}{c|}{Copa}\\
        &Dev&Test\\
        \hline
         FIX&67.8&65.2\\
         SAMPLE&65&64\\
         FIRST&66.6&64.8\\
         LAST&67.2&62.8\\
         UNIFORM&68.8&66.6\\
         Our&\textbf{70.2}&\textbf{67.4}\\
         \toprule
    \end{tabular}
    \caption{The accuracy of different Sampling \\\hspace{2em} strategies}
\label{table_sample_strategy}
  \end{minipage} 
\end{figure*}


In this section, we explore and compare five different sampling strategies to organize the source dataset at MAML steps. 1. Sampling task uniformly (UNIFORM). 2. Using the size of the source dataset as the probability (FIX). 3. Using the softmax probability of the difference $\{\rm D_{j}\}_{j=1}^{k}$ (SAMPLE). 4. Using the top numbers of source based on difference (FIRST). 5. Using the last numbers of source based on difference (LAST). 6. Our reinforcement-based sampling method. All results are showed in the Table \ref{table_sample_strategy}. 
From the table, we can known FIRST has an improvement over LAST and SAMPLE. In the Figure \ref{test_acc}, we show the results of these three sampling strategies. It is obvious that the FIRST have an improvement over the LAST and SAMPLE on the whole curve, indicating the metric can guide the choose of tasks. And our proposed sampling method surpasses all others methods in the final result, which proves that by incorporating the long-term information, we can organize difference sources well and get a better result. 

\section{Conclusion}
In this work, we propose a novel method named reinforcement-based meta transfer for low-resource cross-lingual cross-task commonsense reasoning on low resource setting. Our
method considers multiple sources meta learning
and target lingual adaptation into a unified framework,
which is able to learn a general representation from
multiple sources and alleviate the discrepancy between source and target by using the policy network. We demonstrate the superiority of our methods on both supervised setting
and unsupervised adaptation settings over
the state-of-the-arts. In our future work, we explore to
extend this approach for other low resource tasks
in NLP.

\begin{ack}
we would like to thank the anonymous reviewers for their insightful comments.
\end{ack}

\bibliography{main}
\bibliographystyle{plainnat}

\section*{Checklist}


\begin{enumerate}

\item For all authors...
\begin{enumerate}
  \item Do the main claims made in the abstract and introduction accurately reflect the paper's contributions and scope?
    \answerYes{}
  \item Did you describe the limitations of your work?
    \answerYes{}
  \item Did you discuss any potential negative societal impacts of your work?
    \answerYes{}
  \item Have you read the ethics review guidelines and ensured that your paper conforms to them?
    \answerYes{}
\end{enumerate}

\item If you are including theoretical results...
\begin{enumerate}
  \item Did you state the full set of assumptions of all theoretical results?
    \answerNA{}
	\item Did you include complete proofs of all theoretical results?
    \answerNA{}
\end{enumerate}

\item If you ran experiments...
\begin{enumerate}
  \item Did you include the code, data, and instructions needed to reproduce the main experimental results (either in the supplemental material or as a URL)?
    \answerNA{}
  \item Did you specify all the training details (e.g., data splits, hyperparameters, how they were chosen)?
    \answerYes{}
	\item Did you report error bars (e.g., with respect to the random seed after running experiments multiple times)?
    \answerNo{}
	\item Did you include the total amount of compute and the type of resources used (e.g., type of GPUs, internal cluster, or cloud provider)?
    \answerYes{}
\end{enumerate}

\item If you are using existing assets (e.g., code, data, models) or curating/releasing new assets...
\begin{enumerate}
  \item If your work uses existing assets, did you cite the creators?
    \answerYes{}
  \item Did you mention the license of the assets?
   \answerNA{}
  \item Did you include any new assets either in the supplemental material or as a URL?
    \answerNo{}
  \item Did you discuss whether and how consent was obtained from people whose data you're using/curating?
     \answerNA{}
  \item Did you discuss whether the data you are using/curating contains personally identifiable information or offensive content?
     \answerNA{}
\end{enumerate}

\item If you used crowdsourcing or conducted research with human subjects...
\begin{enumerate}
  \item Did you include the full text of instructions given to participants and screenshots, if applicable?
     \answerNA{}
  \item Did you describe any potential participant risks, with links to Institutional Review Board (IRB) approvals, if applicable?
    \answerNA{}
  \item Did you include the estimated hourly wage paid to participants and the total amount spent on participant compensation?
     \answerNA{}
\end{enumerate}

\end{enumerate}


\appendix

\section{Source datasets}
\begin{itemize}
\setlength{\itemsep}{0pt}
\setlength{\parsep}{0pt}
\setlength{\parskip}{0pt}
    \item SocialIQA \cite{sap-etal-2019-social} is a collection of 38K questions for probing  social situations and interactions. Each question is paired with four candidate answers.
    \item $\rm \alpha$NLI \cite{gabriel2019coopnet} contains over 20K commonsense narrative contexts and 200K explanations. It requires  to identify the best explanation from two candidates.
    \item CosmosQA \cite{huang-etal-2019-cosmos} is a four-choice reading comprehension dataset, focusing on understanding narrative and  interpreting likely causes and effects of events.
    \item CommonsenseQA \cite{talmor-etal-2019-commonsenseqa} contains 12,247 question-answer pairs. Each question has five candidate answers, aiming at  investigating question answering with prior knowledge.
    \item Swag \cite{zellers-etal-2018-swag} requires to reason about  situations and to anticipate what might happen next, with 113K
    four-choice questions.
    \item HellaSwag \cite{zellers-etal-2019-hellaswag} uses pre-trained language models to perform  adversarial filtering  to collect  70K challenging  examples.
\end{itemize}

\section{Experimental settings}
All our experiments were  conducted on a single NVIDIA TiTan 24GB  GPU  with  PyTorch.  We adopted  mBERT (bert-base-multilingual-cased)\footnote{\href{https://github.com/huggingface/transformers}{https://github.com/huggingface/transformers}} as the backbone. The hidden size of the  adapter is 128. Due to the memory  limitation, the maximal  input length of mBERT was set  to 128.

Our implementation of MetaXCR was based on higher\footnote{\href{https://github.com/facebookresearch/higher}{https://github.com/facebookresearch/higher}} \cite{grefenstette2019generalized}.
For CTML and CLML, unless  otherwise  specified,  a  CTML meta-batch  contained   3  tasks while  CLML meta-batch  1 task. The batch size for each task was 12. The base model  and meta-learner  were   both  optimized  with  Adamw \cite{DBLP:conf/iclr/LoshchilovH19} optimizer, and the learning rate was  set to 0.00015 and 0.0002 respectively. The number of inner gradient steps was 1, and  we used accuracy  convergence as a criteria for stopping.

\section{Comparing Cross-lingual adaptation and mono-lingual adaptation}
\begin{figure}[hbpt]
\centering
\includegraphics[scale=0.5]{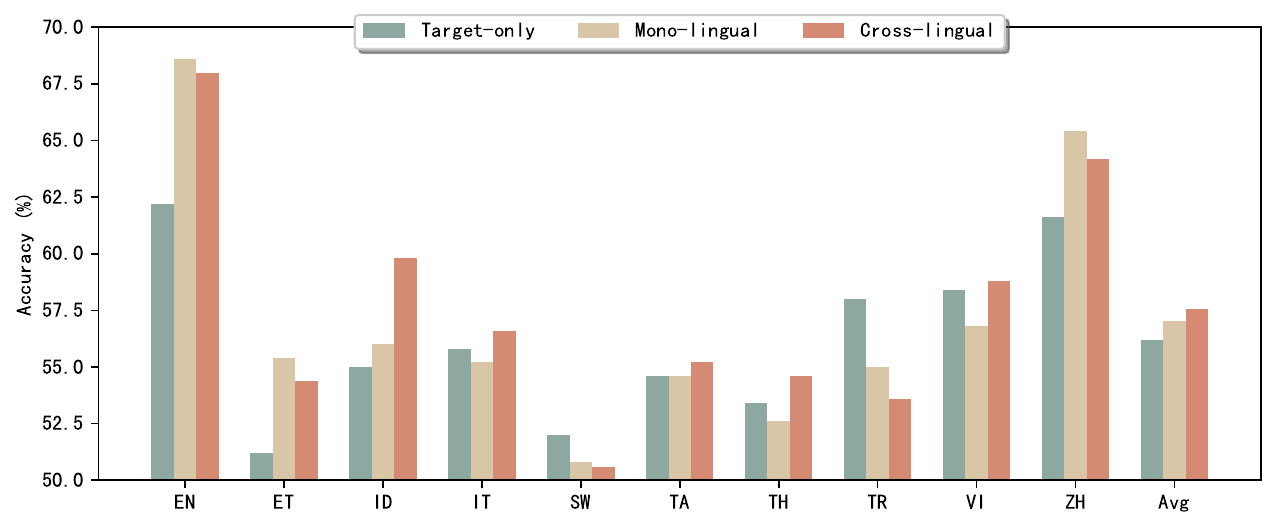}
\caption
{The accuracy results for Target-only, mono-lingual, cross-lingual training method.
}
\label{figure_total_acc}
\end{figure}
In this section, we have constructed two baselines to compare our CLML with other traditional methods. The target-only method performs fine-tuning on target language development sets and mono-lingual meta  learning adaptation uses the target language development set as support task and query task instead of using English data. 

The experimental results are shown in the Figure \ref{figure_total_acc}. We observe that the cross-lingual meta adaptation method performs best in 9 languages, while the target-only method performs the worst. On the one hand, because of direct fine-tune, over-fitting problem tend to occur. On the other hand, due to the existence of the aforementioned anglo-centric bias, the effect of fine-tune is not good. For mono-lingual meta learning adaptation, there is also a certain gap with cross-lingual meta learning adaption, which is in line with our conjecture that the model can benefit more from learning how to transfer English knowledge to other languages.


\end{document}